\documentclass[10pt,twocolumn,letterpaper]{article}

\usepackage{iccv}
\usepackage{times}
\usepackage{epsfig}
\usepackage{graphicx}
\usepackage{amsmath}
\usepackage{amssymb}
\usepackage{multirow}
\usepackage{booktabs}

\usepackage[pagebackref=true,breaklinks=true,letterpaper=true,colorlinks,bookmarks=true]{hyperref}

\iccvfinalcopy 

\def\iccvPaperID{2285} 

\ificcvfinal\fi

\begin{document}

\title{Empowering Low-Light Image Enhancer through Customized Learnable Priors}

\author{
{Naishan Zheng} \textsuperscript{1}\footnotemark[1], 
{Man Zhou} \textsuperscript{1}\footnotemark[1], 
{Yanmeng Dong},  
{Xiangyu Rui} \textsuperscript{2}, 
{Jie Huang} \textsuperscript{1}, 
{Chongyi Li} \textsuperscript{3}, 
{Feng Zhao} \textsuperscript{1}\footnotemark[2] \\
\textsuperscript{1}University of Science and Technology of China \\
\hspace{-0.4cm}\textsuperscript{2}Xi’an Jiaotong University, 
\hspace{0.1cm}\textsuperscript{3}Nankai University \\
\tt\small \{nszheng,manman,hj0117\}@mail.ustc.edu.cn, 
\\ \tt\small
\{dongym.aca,xyrui.aca,lichongyi25\}@gmail.com, fzhao956@ustc.edu.cn\\
}

\maketitle
\ificcvfinal\thispagestyle{empty}\fi

\begin{abstract}
\renewcommand{\thefootnote}{\fnsymbol{footnote}}
\footnotetext[1]{Both authors contributed equally to this research.}
\footnotetext[2]{Corresponding author.}

Deep neural networks have achieved remarkable progress  in enhancing low-light images by improving their brightness and eliminating noise. 
However, most existing methods  construct end-to-end mapping networks heuristically, neglecting the intrinsic prior of image enhancement task and lacking transparency and interpretability. 
Although some unfolding solutions have been proposed to relieve these issues, they rely on proximal operator networks that deliver ambiguous and implicit priors. 
In this work, we propose a paradigm for low-light image enhancement that explores the potential of customized learnable priors to improve the transparency of the deep unfolding paradigm.
Motivated by the powerful feature representation capability of Masked Autoencoder (MAE), we customize MAE-based illumination and noise priors and redevelop them from two perspectives: 
1) \textbf{structure flow}: we train the MAE from a normal-light image to its illumination properties and then embed it into the proximal operator design of the unfolding architecture; and
2) \textbf{optimization flow}: we train MAE from a normal-light image to its gradient representation and then employ it as a regularization term to constrain  noise in the model output. 
These designs improve the interpretability and representation capability of the model.
Extensive experiments on multiple low-light image enhancement datasets demonstrate the superiority of our proposed paradigm over state-of-the-art methods.
Code is available at https://github.com/zheng980629/CUE.
\end{abstract}

\begin{figure}[t]
	\centering
	\includegraphics[width=0.45\textwidth]{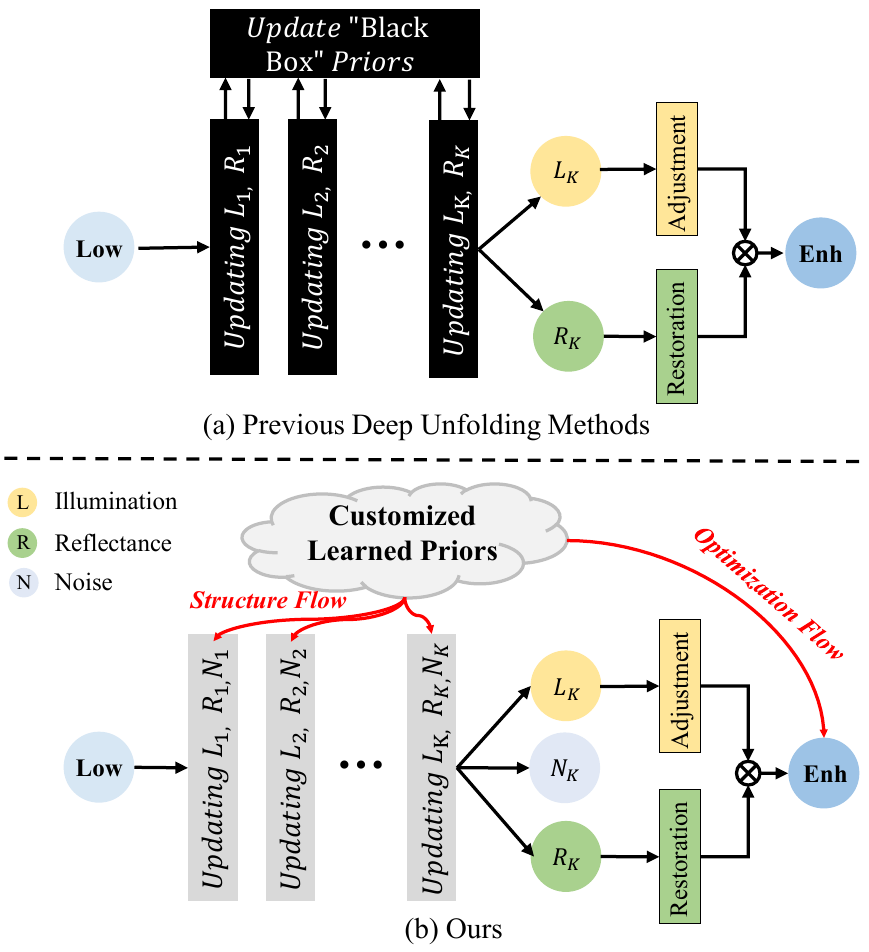}
	\vspace{-0.1cm}
	\caption{Comparison between previous deep unfolding low-light enhancement methods and our proposed paradigm. (a) Previous works deliver the ambiguous and implicit priors by the heuristically proximal networks in a ``black box'' manner; and (b) our CUE explores the potential of customized learnable priors on low-light image enhancement to improve the transparency of the deep unfolding paradigm.}
	\label{fig:teaser}
	\vspace{-0.3cm}
\end{figure}

\section{Introduction}
Low-light conditions often result in images with limited visibility and noise, which can negatively affect downstream computer vision tasks. 
To recover details buried in low-light images and remove noise effects, the field of low-light image enhancement has received significant attention. 
Existing approaches can be classified into two categories:  traditional methods and deep learning-based methods.

Traditional methods for low-light image enhancement involve formulating the problem as an optimization task and using image priors as regularization terms to constrain the solution space.
One such representative method is based on the Retinex theory\cite{land1977retinex}, which assumes that an image can be decomposed into reflectance and illumination components where the former keeps consistency under any lighting conditions and the latter reflects variations in brightness.
Nonetheless,  estimating illumination and reflectance terms simultaneously is challenging. 
To overcome this issue, based on the Retinex theory, well-designed priors of the illumination and reflectance terms are proposed. 
For example, an $\ell_{2}$ norm on illumination gradients is proposed by \cite{SRIE,fu2015probabilistic} to ensure  smoothness but generate blurred borders around areas where the illumination suddenly changes. 
\cite{Roubust_retinex} further introduces a noise term and develops a $\ell_{1}$ prior to constrain illumination gradients, maintaining the overall structure of the illumination map. 
However,  hand-crafted priors are difficult to design and have limited representative ability in complex scenes, hindering their practical usage.

Inspired by the powerful learning capability of deep learning, explosive learning-driven methods have undeniably ushered in a new era for low-light image enhancement~\cite{HPEU,FECNet,ECLNet,LowLightZhang,Huang2023CVPR,Huang2022CVPR,9847530,zheng2022enhancement,liu2022learning,ma2021learning,liu2021underexposed,ma2022low,ma2022pia,jin2021bridging,xue2022best,ma2023bilevel,ma2022practical,liu2022twin}.
For example, RetinexNet~\cite{RetinexNet} and KinD~\cite{KinD} are two popular methods that use deep learning to enhance low-light images. RetinexNet integrates Retinex decomposition and illumination adjustment into an end-to-end trainable network, while KinD separately trains layer decomposition, illumination adjustment, and reflectance restoration subnetworks. 
Despite their success, these methods often construct black-box networks without considering the intrinsic priors of illumination and reflectance components, leading to a lack of transparency and interpretability.

To improve the interpretability, a model-driven deep unfolding paradigm,  URetinexNet~\cite{URetinexNet}, has been proposed.
It formulates the Retinex decomposition as an implicit prior regularized model but neglects the effects of noise pollution. 
Moreover, this method delivers the illumination and reflectance priors in a vague and ambiguous manner through empirically constructed proximal operator networks.
Thus, the inherent properties of these two components are not considered, leading to an ambiguous and implicit prior principle (see Fig.~\ref{fig:teaser}).
We therefore wonder, \emph{``Can we customize learnable priors for illumination and noise terms that leverage their intrinsic properties?''}.

To answer this question, we first analyze the characteristics of the illumination and noise components: 
1) \textbf{illumination prior}: Based on the Retinex theory, the illumination component should be smooth and preserve the structure while adapting to different lighting conditions; 
2) \textbf{noise prior}: In low-light images, noise is inherent and cannot be removed simply by adjusting brightness, \emph{i.e.}, irrelevant to enhanced lightness. 

Motivated by the above analysis, we aim at exploring the potential of customized learnable priors for low-light image enhancement to improve the transparency of the deep unfolding paradigm.
Our proposed method is called Customized Unfolding Enhancer (CUE). 
To achieve this, we utilize the innate feature representation capability of Masked Autoencoder (MAE) to customize MAE-based illumination and noise priors with a masked image modeling strategy. 
Specifically, the  illumination prior is trained from a normal-light image to its illumination map filtered by a bilateral filter, reducing noise without altering the intrinsic structure~\cite{bilateralGrid}. 
The noise prior aims to learn the histograms of oriented gradients of a normal-light image, which presents the gradient variation while being irrelevant to enhanced lightness~\cite{hog}. 
To integrate the customized priors into the low-light image enhancement process, we redevelop the two learned priors from two perspectives:
1) \textbf{structure flow}: embedding the learned illumination prior into the design of the proximal operator in the Retinex decomposition unfolding process; and 
2) \textbf{optimization flow}: redeveloping the learned noise prior as a regularization term to  eliminate noise by minimizing the gradient presentation difference between the enhanced and normal-light images. 
Extensive experimental results demonstrate the superiority of our paradigm over state-of-the-art methods. 
Additionally, we also verify the effectiveness of the proposed learnable noise prior for image denoising.

Our contributions are summarized as follows:
\begin{itemize}
    \item We activate the potential of customized learnable illumination and noise priors via a new deep unfolding paradigm for low-light image enhancement.
    \item From the structure flow, we embed the MAE-based customized illumination prior into the unfolding architecture to improve the transparency and interpretability of the unfolding structure.
    \item From the optimization flow, we redevelop the MAE-based customized noise prior as a regularization term to constrain the gradient representation consistency.
    \item  Our experiments on multiple low-light image benchmarks show that the proposed paradigm outperforms state-of-the-art methods, and our customized noise prior is effective for image denoising.
\end{itemize}

\section{Related Work}
\subsection{Low-Light Image Enhancement}
There are three groups of traditional image enhancement techniques: Histogram Equalization (HE), Gamma Correction (GC), and Retinex theory~\cite{land1977retinex}.
HE-based techniques stretch the dynamic range of the low-light images with various complicated priors~\cite{dynamicHistogram,histogramModification,contrastEnhance,weightedHE}.
GC-based~\cite{gamma,variationalGamma} methods expand the dynamic range of an image by applying an exponential function to each pixel individually. 
In addition, Retinex-based variants~\cite{SRIE,Roubust_retinex,NPE,Roubust_retinex} have developed priors to constrain the solution space of the illumination and reflectance maps.
However, the inflexible hand-crafted priors may not generalize well in complex scenes.

\begin{figure*}[!t]
	\centering
	\includegraphics[width=0.93\textwidth]{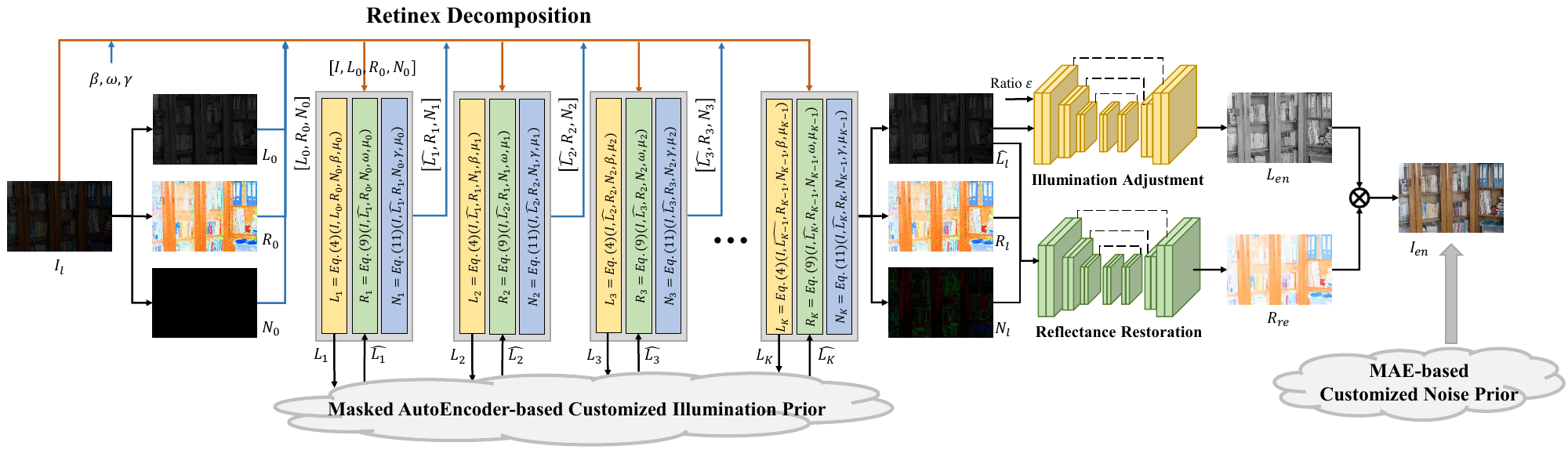}
	\caption{The end-to-end training paradigm of the proposed Customized Unfolding Enhancer (CUE) includes an unfolding Retinex decomposition step embedded with a Masked Autoencoder-based customized illumination prior (see Fig.~\ref{fig:BilateralGrid}) and an enhancement step. First, a low-light image $\mathbf{I_{l}}$ is decomposed into illumination, reflectance, and noise terms, $[\mathbf{L}_{l}, \mathbf{R}_{l}, \mathbf{N}_{l}]$, by the unfolding Recinx decomposition step. Then, the illumination adjustment and reflectance restoration networks are employed to enhance the illumination and restore the reflectance components. Finally, the MAE-based noise prior (see Fig.~\ref{fig:hog}) is redeveloped as a regularization term to further eliminate noise in the enhanced image by constraining the gradient representation.}
\vspace{-0.3cm}
\label{fig:framework}
\end{figure*}

In the past decade, deep learning-based low-light image enhancement approaches~\cite{LLNet,RetinexNet,KinD,DSLR,DRBN,TBEFN,semantic_retinex,LPNet,progressive_retinex,RetinexDIP,ZERO_DCE++,zhuang2021bayesian,li2021low,zhou2022lednet,li2023embedding,zhou2023fourmer}, have achieved  remarkable performance gains. 
Wei \emph{et al.}~\cite{RetinexNet} first attempted to integrate Retinex decomposition and illumination adjustment into an end-to-end network.
Zhang \emph{et al.}~\cite{KinD} employed three subnetworks for decomposition, reflection restoration, and illumination adjustment. 
Wang \emph{et al.}~\cite{DeepUPE} proposed to enhance the underexposed images by estimating an illumination map. 
Guo \emph{et al.}~\cite{ZERO_DCE} adjusted the dynamic range of the low-light images by predicting pixel-wise mapping without paired or unpaired data. 
However, these learning-based methods empirically build network architectures in a black-box fashion, neglecting the inherent prior and lacking adequate interpretability. 
To address this limitation, Liu \emph{et al.}~\cite{RUAS} developed a Retinex-inspired unrolling strategy to discover the architectures embedded with atomic priors automatically.
Wu \emph{et al.}~\cite{URetinexNet} proposed to unroll the Retinex decomposition and formulated it as an implicit prior regularized model.
Zheng \emph{et al.}~\cite{TVUnfolding} unfolded the total variant minimization algorithms to provide fidelity and smoothness constraints via a learnable noise level map. 
While these deep unfolding solutions improve the interpretability, the priors are still vaguely delivered by convolutional neural networks constructed empirically in a black-box manner, resulting in an ambiguous and implicit prior principle.

\subsection{Masked Image Modeling}
Inspired by the success of BERT~\cite{bert} in NLP, BEiT~\cite{beit} proposes a masked image modeling technique that predicts the visual token of each masked patch as a pretext task.
MAE~\cite{MAE} presents an asymmetric encoder-decoder paradigm that focuses on the visible patch. The encoder operates on this patch, while the decoder reconstructs the original image from the latent representation and mask tokens.
To improve representation capability, CAE~\cite{CAE} decouples the encoding and pretext task completion roles through an alignment constraint and a latent contextual regressor. 
MVP~\cite{MVP} aims to create robust and multi-modal representation by predicting the guidance from other modalities. 
However, their primary purpose is to serve as a pre-training technique to enhance performance on downstream tasks.

\section{Methodology}
In this section, we introduce the proposed Customized Unfolding Enhancer (CUE) paradigm, which consists of a structure flow and an optimization flow, as illustrated in Fig.~\ref{fig:framework}. 
In the structure flow, we first describe an unfolding Retinex decomposition step embedded with an MAE-based customized illumination prior and provide an enhancement step. 
Then we go into the optimization flow, including Retinex decomposition, enhancement, and a gradient representation regularization derived from an MAE-based customized noise prior.

\subsection{Structure Flow}
\subsubsection{Retinex Decomposition}
Classical Retinex theory assumes that an observed image can be decomposed into illumination and reflectance components. Due to the inevitable noise in low-light images, \cite{Roubust_retinex} introduces a noise term:
\begin{equation}
    \mathbf{I} = \mathbf{R} \circ \mathbf{L} + \mathbf{N}, 
    \label{imaging}
\end{equation}
where $\rm I$, $\rm R$, $\rm L$, and $\rm N$ denote the observed image, reflectance, illumination, and noise, respectively. The operator $\circ$ represents the element-wise multiplication. Referring to the above observation model, the reflectance, illumination, and noise terms can be simultaneously obtained by solving the minimization problem: 
\begin{subequations}
\begin{equation}
\arg \min _{\mathbf{R}, \mathbf{L}, \mathbf{N}}\|\mathbf{N}\|_{1} + \gamma \rho_{1}(\mathbf{N})+\beta \rho_{2}(\mathbf{L})+\omega \rho_{3}(\mathbf{R}), 
\end{equation}
\begin{equation}
s.t.  \mathbf{I}=\mathbf{R} \circ \mathbf{L}+\mathbf{N}, 
\end{equation}
\label{optimization}
\end{subequations}
where $\|\cdot\|_{1}$ denotes $\ell_{1}$ norm, $\rho_{1}$, $\rho_{2}$, and $\rho_{3}$ are the regularization terms denoting imposed priors over $\mathbf{N}$, $\mathbf{L}$, and $\mathbf{R}$, and $\gamma$, $\beta$, and $\omega$ are trade-off parameters. $\|\mathbf{N}\|_{1}$ is a general prior that simply constrains the noise's sparsity, which cannot accurately model the noise distribution in the low-light image. 
Therefore, we incorporate an implicit noise prior, $\rho_{1}(\mathbf{N})$, to further estimate the extreme noise.

By introducing a penalty function to remove the equalization constraint, Eq.~\eqref{optimization} is written as: 
\begin{equation}
\begin{aligned}
    \mathcal{L}(\mathbf{N}, \mathbf{L}, \mathbf{R}) &= \|\mathbf{N}\|_{1} + \gamma \rho_{1}(\mathbf{N})+\beta \rho_{2}(\mathbf{L}) \\
    &+ \omega \rho_{3}(\mathbf{R}) +\frac{\mu}{2}\|\mathbf{I}-\mathbf{R} \circ \mathbf{L}-\mathbf{N}\|_{F}^{2},
\end{aligned}
\label{penalty}
\end{equation}
where $\|\cdot\|_{F}$ represents Frobenius norm and $\mu$ is a penalty parameter. 
The equivalent objective function can be solved by iteratively updating $\mathbf{L}$, $\mathbf{R}$, and $\mathbf{N}$ while considering other variables that have been estimated in
previous iterations as constants. 
Here we present the sub-problem solutions for the $(k+1)$-th iteration.

\textbf{L sub-problem:}
Given the estimated reflectance and noise at iteration $k$, $\mathbf{R}_{k}$, and $\mathbf{N}_{k}$, $\mathbf{L}$ can be updated as:

\begin{equation}
\begin{aligned}
\mathbf{L}_{k+1} &=\arg \min _{\mathbf{L}} \mathcal{L}\left(\mathbf{N}_{k}, \mathbf{L}, \mathbf{R}_{k}\right) \\
&=\arg \min _{\mathbf{L}} \beta \rho_{2}(\mathbf{L})+\underbrace{\frac{\mu_{k}}{2}\left\|\mathbf{I}-\mathbf{R}_{k} \circ \mathbf{L}-\mathbf{N}_{k}\right\|_{F}^{2}}_{g(\mathbf{L})}.
\end{aligned}
\label{L-sub}
\end{equation}

By applying the proximal gradient method to Eq.~\eqref{L-sub}, we can derive:
\begin{equation}
\begin{aligned}
\mathbf{L}_{k+1} &= \operatorname{prox}_{\beta \rho_{2}}\left(\mathbf{L}_{k}-\alpha_{1} \nabla g\left(\mathbf{L}_{k}\right)\right),
\end{aligned}
\label{proxL}
\end{equation}
where $\operatorname{prox}_{\beta \rho_{2}}$ is the proximal gradient operator corresponding to the prior $\rho_{2}$, $\alpha_{1}$ denotes the updating step size, and $\nabla g\left(\mathbf{L}_{k}\right) = \mu_{k} \mathbf{R}_{k} \circ\left(\mathbf{R}_{k} \circ \mathbf{L}_{k}+\mathbf{N}_{k}-\mathbf{I}\right)$.

\begin{figure}[t]
	\centering
	\includegraphics[width=0.45\textwidth]{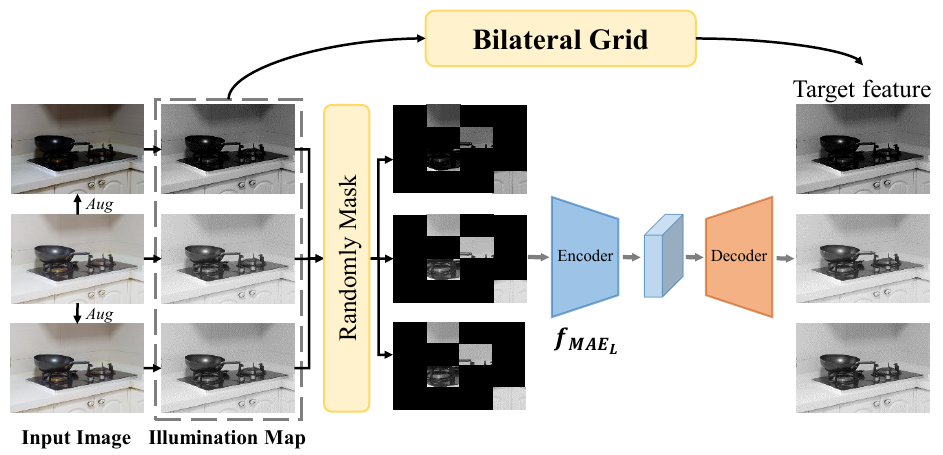}
	\caption{The MAE-based customized illumination prior. A UNet-like convolution neural network with a mask image modeling strategy takes the whole randomly masked illumination map as input to predict the corresponding map processed by a bilateral filter. After pre-training, the encoder $f_{MAE_{l}}$ is embedded into the design of the proximal operation in the L sub-problem.}
	\label{fig:BilateralGrid}
	\vspace{-0.2cm}
\end{figure}

Existing methods for designing the functional form of $\operatorname{prox}_{\beta \rho{2}}$ construct implicit networks in a black box manner, resulting in an ambiguous and implicit prior principle. 
However, according to Retinex theory, the illumination term should be smooth while preserving the structure and adapting to the diverse lighting conditions~\cite{land1977retinex}.
Fortunately, the bilateral filter operator is inherently  capable of  filtering noise while preserving edges~\cite{bilateralGrid}. 
In this paper, standing on the shoulder of MAE~\cite{MAE}, which is trained using the masked image modeling strategy and embraces powerful feature representation capability, we design a pre-trained MAE with the illumination map of a normal-light image filtered by the bilateral filter as the target feature, thereby creating a customized illumination prior.
We then embed the learned illumination prior into the design of $\operatorname{prox}_{\beta \rho{2}}$ to enhance its transparency.

The original MAE is an asymmetric encoder-decoder architecture that reconstructs the masked pixels given partially visible patches. 
During pre-training, a high proportion of the input image is masked, and the encoder only maps the visible patches into a latent feature representation. 
Then, a lightweight decoder reconstructs the original images in pixels from the latent representation and mask tokens. 
After the pre-training, the decoder is discarded, and the encoder is applied to uncorrupted images (full sets of patches) for feature representation.

\begin{figure}[t]
	\centering
	\includegraphics[width=0.35\textwidth]{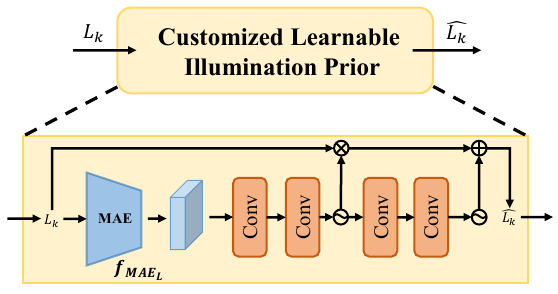}
	\vspace{-0.1cm}
	\caption{The details of the design of $\operatorname{prox}_{\beta \rho_{2}}$.}
	\label{fig:prox}
	\vspace{-0.4cm}
\end{figure}

In this paper, as shown in Fig.~\ref{fig:BilateralGrid}, deviated from the original MAE, we implement a UNet-like convolution neural network with a customized target feature in the masked image modeling strategy:
\begin{itemize}
    \item employing the vanilla convolution operator to construct the encoder-decoder architecture; 
    \item implementing diverse lighting conditions by augmenting the input image with gamma transformation; 
    \item dividing the corresponding illumination maps (obtained by the max intensity of RGB channels) into regular non-overlapping regions, randomly sampling a subset of regions, and masking the remaining ones while maintaining the whole structure of the map; 
    \item processing all the regions (both visible and masked regions) through the encoder and decoder to reconstruct the illumination map filtered by the bilateral filter; 
    \item for accentuation, the input of the network is the complete image structure rather than image patches.
\end{itemize}

After pre-training, we incorporate its encoder $f_{{MAE}_{L}}$ into the design of $\operatorname{prox}_{\beta \rho_{2}}$ as described in Fig.~\ref{fig:prox}. 
Specifically, feeding the $(k+1)$-th iteration $\mathbf{L}_{k+1}$ into $f_{{MAE}_{L}}$ to generate the feature representation possessing the properties of the illumination map:
\begin{equation}
    \mathbf{L}_{fp} = f_{{MAE}_{L}}\left(\mathbf{L}_{k+1}\right).
\end{equation}
Then, based on the prior features, an illumination map, $\hat{\mathbf{L}}_{k+1}$, integrated with the customized illumination prior is formulated as:
\begin{equation}
    \hat{\mathbf{L}}_{k+1} = \mathbf{L}_{k+1} * Sig(Conv(\mathbf{L}_{fp}\uparrow)) + Sig(Conv(\mathbf{L}_{fp}\uparrow)),
\end{equation}
where $\uparrow$ denotes upsampling to the same spatial size as $\hat{\mathbf{L}}_{k+1}$, $Conv$ indicates three convolution layers to generate the modulation parameter, and $Sig$ is the Sigmoid function. 
The qualitative and quantitative evaluations of the effectiveness of the $f_{{MAE}_{L}}$ is presented in Fig.~\ref{fig:ablation} and Table~\ref{tab:component}.

\textbf{R sub-problem:}
Dropping the terms unrelated to $\mathbf{R}$ gives the following optimization problem:
\begin{equation}
\begin{aligned}
\mathbf{R}_{k+1} &=\arg \min _{\mathbf{R}} \mathcal{L}\left(\mathbf{N}_{k}, \hat{\mathbf{L}}_{k+1}, \mathbf{R}\right) \\
&=\arg \min _{\mathbf{R}} \omega \rho_{3}(\mathbf{R})+\underbrace{\frac{\mu_{k}}{2}\left\|\mathbf{I}-\mathbf{R} \circ \hat{\mathbf{L}}_{k+1}-\mathbf{N}_{k}\right\|_{F}^{2}}_{h(\mathbf{R})}.
\end{aligned}
\label{R-sub}
\end{equation}
Similarly, Eq.~\eqref{R-sub} is written as:
\begin{equation}
    \mathbf{R}_{k+1} = \operatorname{prox}_{\omega \rho_{3}}\left(\mathbf{R}_{k}-\alpha_{2} \nabla h\left(\mathbf{R}_{k}\right)\right),
\end{equation}
where $\operatorname{prox}_{\omega \rho_{3}}$ is the proximal gradient operator corresponding to the prior $\rho_{3}$, implemented by two Conv layers followed by the ReLU activation, $\alpha_{2}$ indicates the updating step size, and $\nabla h\left(\mathbf{R}_{k}\right)=\mu_{k} \hat{\mathbf{L}}_{k+1} \circ\left(\hat{\mathbf{L}}_{k+1} \circ \mathbf{R}_{k}+\mathbf{N}_{k}-\mathbf{I}\right)$.

\textbf{N sub-problem:}
Collecting the terms related to $\mathbf{N}$ leads to the problem as follows:
\begin{equation}
\begin{aligned}
\mathbf{N}_{k+1} &=\arg \min _{\mathbf{N}} \mathcal{L}\left(\mathbf{N}, \hat{\mathbf{L}}_{k+1}, \mathbf{R}_{k+1}\right) \\
&=\arg \min _{\mathbf{N}} \left\|\mathbf{N}\right\|_{1} + \gamma \rho_{1}(\mathbf{N}) \\
&+\frac{\mu_{k}}{2}\left\|\mathbf{I}-\mathbf{R}_{k+1} \circ \hat{\mathbf{L}}_{k+1}-\mathbf{N}\right\|_{F}^{2}.
\end{aligned}
\label{N-sub}
\end{equation}
The solution of Eq.~\eqref{N-sub} is formulated as:
\begin{equation}
    \mathbf{N}_{k+1} = \operatorname{prox}_{\gamma \rho_{1}}(\rm Shrink(\mathbf{I} - \mathbf{R}_{k+1} \circ \hat{\mathbf{L}}_{k+1}, \frac{1}{\mu_{k}})),
\end{equation}
where $\rm Shrink(\mathbf{X}, \eta) = max\left\{|\mathbf{X}| - \eta, 0\right\} \cdot \rm sign(\mathbf{X})$, $\rm sign(\cdot)$ denotes the sign function, and 
$\operatorname{prox}_{\gamma \rho_{1}}$ indicates the proximal gradient operator corresponding to the prior $\rho_{1}$, achieved by two Conv layers followed by the ReLU 
activation. 

\begin{figure}[!t]
	\centering
	\includegraphics[width=0.45\textwidth]{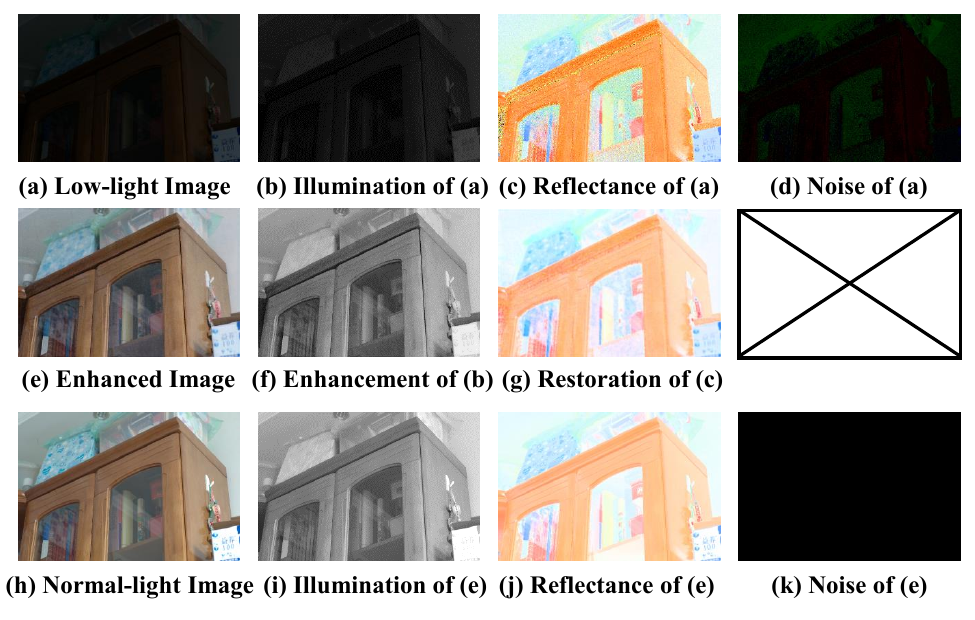}
	\vspace{-0.1cm}
	\caption{The first and third rows present decomposed illumination, reflectance, and noise components of the low-/normal- light images, [$\mathbf{L}_{l}$, $\mathbf{R}_{l}$, $\mathbf{N}_{l}$] and [$\mathbf{L_n}$, $\mathbf{R}_{n}$, $\mathbf{N}_{n}$], respectively, where $\mathbf{N}_{n}$ is equal to zero. The second row illustrates the enhanced image, the enhanced illumination map, and the restored reflectance map, [$\mathbf{I}_{en}, \mathbf{L}_{en}, \mathbf{R}_{re}$].}
	\label{fig:decomposition}
	\vspace{-0.2cm}
\end{figure}

Fig.~\ref{fig:decomposition} presents the decomposed illumination, reflectance, and noise terms of the paired low-/normal- light images.

\subsubsection{Enhancement}
After the Retinex decomposition, the low-light and normal-light images, $I_{l}$ and $I_{n}$, are decomposed into their illumination, reflectance, and noise terms, \emph{i.e.}, [$\mathbf{L}_{l}$, $\mathbf{R}_{l}$, $\mathbf{N}_{l}$] and [$\mathbf{L_n}$, $\mathbf{R}_{n}$, $\mathbf{N}_{n}$], where $\mathbf{N}_{n}$ is equal to zero.

Due to the absence of an optimal light level for images, a flexible illumination adjustment system is required. 
Following~\cite{KinD}, an indicator with the value $\varepsilon$ is combined with $\mathbf{L}_{l}$ as the input of the illumination adjustment network $(\varepsilon=mean(\mathbf{L}_{n} / \mathbf{L}_{l})$ for training and specified by users for inference). 
As shown in Fig.~\ref{fig:framework}, the illumination adjustment network is implemented by a U-Net~\cite{unet} with three scales of 12, 24, and 48 channels, respectively. 
Furthermore, the reflectance maps of the low-light images are degraded by color deviation and noise pollution~\cite{KinD}, which are significantly associated with illumination and noise distribution. 
Thus, we integrate the illumination and noise maps, $\mathbf{L}_{l}$ and $\mathbf{N}_{l}$, with the degraded reflectance, $\mathbf{R}_{l}$, into the reflectance restoration network. 
Its structure is the same as the illumination adjustment network. 
Finally, the enhanced image, $\mathbf{I}_{en}$, will be obtained by element-wise multiplication between the enhanced illumination and the restored reflectance maps: $\mathbf{I}_{en} = \mathbf{L}_{en} \circ \mathbf{R}_{re}$. 
Fig.~\ref{fig:decomposition} presents the enhanced illumination, restored reflectance components, and the corresponding enhanced image.

\subsection{Optimization Flow}
Orthogonal to structure design, we introduce elaborately designed loss functions to enable the network for better optimization, thus producing pleasing images.

\textbf{Retinex decomposition:}
Recovering illumination, reflectance, and noise terms simultaneously is an ill-posed problem. 
Fortunately, paired low-/normal- light images are available, and the reflectance of a certain scene should be shared across varying light conditions. 
Therefore, we maximize similarity between the decomposed reflectance pair, \emph{i.e.}, $\mathcal{L}_{r s}^{D}=\left\|\mathbf{R}_{l}-\mathbf{R}_{n}\right\|_{1}$. 
Furthermore, the illumination maps should be piece-wise smooth and mutually consistent. 
The illumination smoothness~\cite{KinD} is implemented by $\mathcal{L}_{i s}^{D}=\left\|\frac{\nabla \mathbf{L}_{l}}{\max \left(\left|\nabla \mathbf{I}_{l}\right|, \epsilon\right)}\right\|_{1}+\left\|\frac{\nabla \mathbf{L}_{n}}{\max \left(\left|\nabla \mathbf{I}_{n}\right|, \epsilon\right)}\right\|_{1}$, 
where $\nabla$ denotes the first order derivative operator in horizontal and vertical directions, $|\cdot|$ means the absolute value operator, and $\epsilon$ is set to $0.01$ to avoid zero denominators.
The mutual consistency~\cite{KinD++} is formulated as $\mathcal{L}_{m c}^{D}=\|\mathbf{M} \circ \exp (-c \cdot \mathbf{M})\|_{1}$ with $\mathbf{M}=\left|\nabla \mathbf{L}_{l}\right|+\left|\nabla \mathbf{L}_{n}\right|$. 
It encourages the strong mutual edges to be well preserved and all
weak edges to be effectively suppressed. 
In addition, the reconstruction error is constrained by $\mathcal{L}_{r e}^{D}=\left\|\mathbf{I}_{l}-\mathbf{R}_{l} \circ \mathbf{L}_{l} - \mathbf{N}_{l}\right\|_{1}+\left\|\mathbf{I}_{n}-\mathbf{R}_{n} \circ \mathbf{L}_{n}\right\|_{1}$. In total, the loss function of Retinex decomposition is as follows: 
\begin{equation}
    \mathcal{L}^{D} = \mathcal{L}_{re}^{D} + 0.009 * \mathcal{L}_{rs}^{D} + 0.15 * \mathcal{L}_{ms}^{D} + 0.2 * \mathcal{L}_{is}^{D}.
\end{equation}

\textbf{Enhancement:}
The loss function in this step comprises illumination adjustment and reflectance restoration:
\begin{equation}
\begin{aligned}
  \mathcal{L}^{E} &= \left\|\mathbf{R}_{re}-\mathbf{R}_{n}\right\|_{1} + (1- \rm SSIM(\mathbf{R}_{re}, \mathbf{R}_{n})) \\
  &+ \left\|\mathbf{L}_{en}-\mathbf{L}_{n}\right\|_{1} +  \left\|\nabla\mathbf{L}_{en}-\nabla\mathbf{L}_{n}\right\|_{1},
\end{aligned}
\end{equation}
where $\rm SSIM(\cdot, \cdot)$ is the structure similarity measurement.

\begin{figure}[t]
	\centering
	\includegraphics[width=0.45\textwidth]{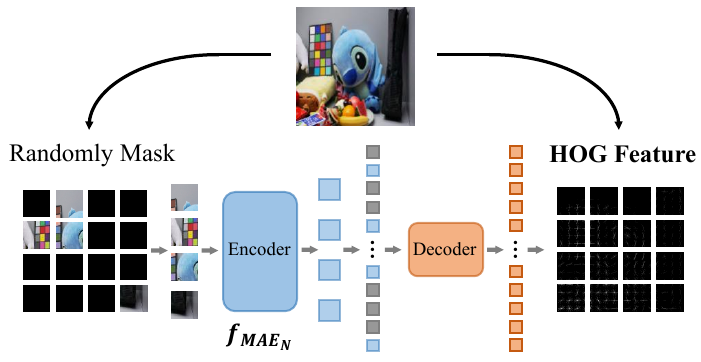}
	\caption{The MAE-based customized noise prior. An MAE-like encoder-decoder architecture takes the visible patches as input and predicts the HOG feature of the masked patches. After pre-training, the encoder $f_{MAE_{N}}$ possesses the powerful gradient representation capability, thus employed as a regularization term to constrain the underlying noise in the enhanced image.}
	\label{fig:hog}
	\vspace{-0.3cm}
\end{figure}

\textbf{Customized noise prior: }
Low-light image enhancement focuses on correcting the luminance while eliminating the amplified noise after enhancement. 
Without noise removal operation, the inherent noise in low-light images will not be eliminated regardless of the adjusted lightness, \emph{i.e.}, being irrelevant to enhanced lightness. 
Histograms of Oriented Gradients~\cite{hog} (HOG) describes the distribution of gradient orientations and magnitudes within a local sub-region while being innately irrelevant to enhanced lightness owing to the local normalization operation. 
Therefore, as shown in Fig.~\ref{fig:hog}, we customize a noise prior with the MAE, where the target feature is the HOG feature of normal-light images:

\begin{itemize}
    \item obtaining the HOG feature of the whole image;
    \item dividing the image into regular non-overlapping patches, randomly sampling a subset of patches, and masking the remaining ones;
    \item flattening the histograms of masked patches and concatenating them into a 1-D vector as the target feature;
    \item the encoder and decoder are implemented by vanilla vision transformers as in MAE.
\end{itemize}

After pre-training, the encoder of the noise prior, $f_{MAE_{N}}$, possesses the powerful gradient representation capability. 
Thus, we redevelop it as a regularization term to suppress noise by imposing gradient representation consistency constraint between the enhanced and normal-light images, which is formulated as:
\begin{equation}
    \mathcal{L}^{N} = \left\| f_{{MAE}_{N}}(\mathbf{I}_{en}) - f_{{MAE}_{N}}(\mathbf{I}_{n}) \right\|_{1}.
\end{equation}
The improvement of $f_{{MAE}_{N}}$ in the noise elimination is illustrated in Fig.~\ref{fig:ablation} and Table~\ref{tab:component}.

In total, the loss function of the proposed CUE for the end-to-end training paradigm is formulated as:
\begin{equation}
    \mathcal{L} = \mathcal{L}^{D} + \mathcal{L}^{E} + \mathcal{L}^{N}.
\end{equation}

\section{Experiments}
\subsection{Experimental Settings}
\textbf{Datasets.} To evaluate the performance of our proposed CUE, we conduct experiments on two commonly used benchmarks, including LOL~\cite{RetinexNet} and Huawei~\cite{huawei}. 
LOL dataset contains 500 paired low-/normal images, and we split 485 for training and 15 for testing as the official selection. 
For the Huawei dataset, we randomly select 2200 images for training and the remaining 280 for testing.

\textbf{Implementations.}
We implement our proposed method with PyTorch on a single NVIDIA GTX 3090 GPU. 
We employ the Adam optimizer with $\beta_{1}=0.9$, $\beta_{2}=0.99$ to update our model for 3000$K$ iterations with a batch of 16.
The initial learning rate is set to $1 \times 10^{-4}$ and decreases with a factor of 0.5 every 50$K$ iterations.
The patch size is set to $64 \times 64$.
The coefficients in loss functions are consistent for training on both LOL and Huawei datasets.

\begin{table*}[t]

\caption{Quantitative comparison on the LOL and Huawei datasets in terms of PSNR$\uparrow$, SSIM$\uparrow$, and NIQE$\downarrow$. The best and the second results are marked in bold and underlined, respectively.}
\centering
\setlength{\tabcolsep}{1.1mm}{
\begin{tabular}{cc|ccccccc}
\toprule
\multirow{2}{*}{Dataset} & \multirow{2}{*}{Metrics} & \multicolumn{7}{c}{Method} \\
\cline{3-9}
    && SRIE & RetinexNet & MBLLEN & GLADNet & TBEFN & KinD & DRBN\\
\midrule
\multirow{2}{*}{LOL/Huawei} 
    & PSNR & 12.28/13.40 & 16.77/16.65 & 17.56/16.63 & 19.72/17.76 & 17.35/16.88 & 20.87/16.48 & 18.65/18.46\\
    & SSIM & 0.596/0.477 & 0.425/0.485 & 0.729/0.526 & 0.680/0.521 & 0.781/0.575 & 0.802/0.540 & 0.801/0.635\\
    & NIQE & 7.506/6.727 & 8.878/7.548 & 3.986/5.418 & 6.475/5.276 & 3.837/5.024 & 4.710/5.704 & 4.454/5.378\\
\midrule
\midrule
\multirow{2}{*}{Dataset} & \multirow{2}{*}{Metrics} & \multicolumn{7}{c}{Method} \\
\cline{3-9}
    && URetinexNet & ZeroDCE++ & EnGAN & RUAS & SCI & CUE (Ours) & CUE++ (Ours)\\
\midrule
\multirow{3}{*}{LOL/Huawei} 
    & PSNR & 21.33/19.25 & 15.53/16.03 & 17.48/17.03 & 16.41/13.76 & 14.78/15.77 & \underline{21.67}/\underline{20.31} & \textbf{21.86}/\textbf{20.38}\\
    & SSIM & \underline{0.834}/0.608 & 0.567/0.507 & 0.674/0.514 & 0.500/0.516 & 0.522/0.480 & 0.774/\underline{0.658} & \textbf{0.841}/\textbf{0.670}\\
    & NIQE & 4.262/5.761 & 7.793/6.641 & 4.154/5.148 & 6.340/6.116 & 7.877/6.879 & \underline{3.776}/\underline{4.908} & \textbf{3.772}/\textbf{4.795}\\
\bottomrule
\end{tabular}}
\vspace{-0.2cm}
\label{tab:LOLHuawei}
\end{table*}

\begin{figure*}[t]
	\centering
	\includegraphics[width=0.92\textwidth]{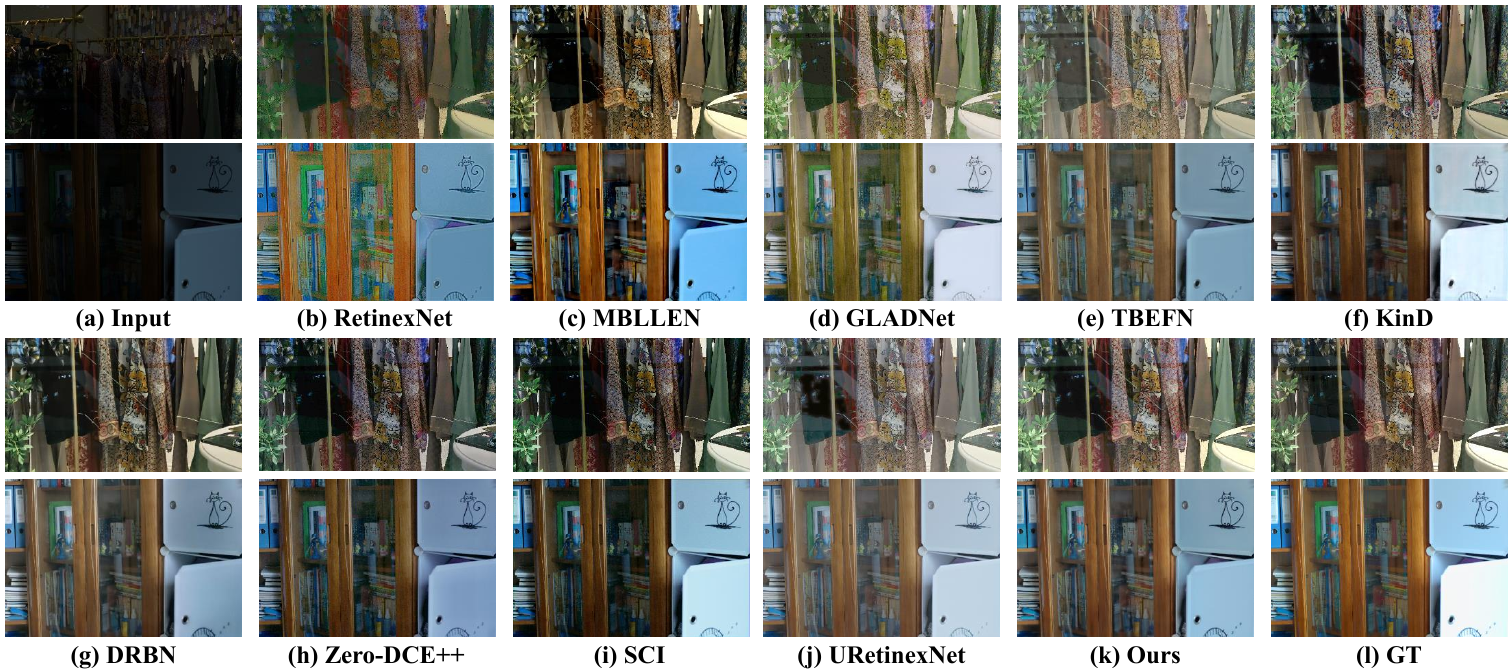}
	\caption{Visual comparison with state-of-the-art methods on the LOL and Huawei dataset. Please zoom in for details.}
	\label{fig:vis}
	\vspace{-0.3cm}
\end{figure*}

\subsection{Comparison with State-of-the-Arts}
We compare our CUE with 12 state-of-the-art approaches: SRIE~\cite{SRIE}, RetinexNet~\cite{RetinexNet}, MBLLEN~\cite{MBLLEN}, GLADNet~\cite{GLADNet}, TBEFN~\cite{TBEFN}, KinD~\cite{KinD}, DRBN~\cite{DRBN}, URetinexNet~\cite{URetinexNet}, Zero-DCE++~\cite{ZERO_DCE++},  EnlightenGAN~\cite{RetinexDIP}, RUAS~\cite{RUAS}, SCI~\cite{SCI} on the LOL and Huawei datasets.

\textbf{Quantitative results.} 
We employ the Peak Signal-to-Noise Ratio (PSNR), Structural Similarity Index (SSIM), and NIQE~\cite{NIQE} as numerical evaluation metrics. 
Note that the URetinexNet~\cite{URetinexNet} and KinD~\cite{KinD} are not trained end-to-end, where decomposition, unfolding, and enhancement stages are trained separately. 
For a fair comparison, we also implement our CUE by individually training the Retinex decomposition, reflectance restoration, and illumination adjustment stages and denote it as CUE++.
To emphasize, CUE++ has the same model size, FLOPs, and runtime as CUE, and the only difference is the training strategy.
As illustrated in Table~\ref{tab:LOLHuawei}, CUE/CUE++ achieves the best results on the two benchmarks, substantiating its effectiveness.

\textbf{Qualitative results.}
Fig.~\ref{fig:vis} shows the subjective visual quality comparison on the challenging extremely low-light images from the Huawei dataset. 
TBEFN~\cite{TBEFN}, GLADNet~\cite{GLADNet}, and URetinexNet~\cite{URetinexNet} generate the results with overexposure, while MBLLEN~\cite{MBLLEN}, SCI~\cite{SCI}, and Zero-DCE++~\cite{ZERO_DCE++} yield underexposed enhanced results. 
Undeniably, DRBN~\cite{DRBN} and KinD~\cite{KinD} predict well-exposed images. 
However, the color distortion in their results is non-ignorable. 
RetinexNet~\cite{RetinexNet}, SCI~\cite{SCI}, and TBEFN~\cite{TBEFN} produce the result with color deviation and artifacts. 
In contrast, our CUE effectively enhances lightness and reveals details while eliminating noise.

\begin{table*}[!t]
\caption{Quantitative comparison in terms of Parameters, FLOPs, and Runtime.}
\begin{center}
\begin{center}
	\setlength{\tabcolsep}{1.3mm}{
		\begin{tabular}{cccccccccccc}
			\toprule 
			Method & RetinexNet & MBLLEN & GLADNet & TBEFN & KinD & DRBN & URetinexNet & EnGAN & SCI & CUE \\
			\midrule
			Params & 0.838M & 0.450M & 1.128M & 0.486M & 8.540M & 0.577M & 0.360M & 8.367M & \textbf{384} &  \underline{0.25M} \\
			FLOPs (G) & 148.54 & 21.37 & 275.32 & \underline{26.33} & 36.57 & 42.41 & 233.09 & 72.61 & \textbf{0.14} & 157.32 \\
			Runtime (s) & 0.023 & 0.159 & 0.024 & 0.035 & 0.068 & 0.140 & 0.115 & \underline{0.011} & \textbf{0.001} & 0.104 \\
			\bottomrule
	\end{tabular}}
\end{center}
\end{center}
\vspace{-0.5cm}
\label{tab:runtime}
\end{table*}

\textbf{Computational complexity.}
We analyze the computational complexity of baselines and our CUE.
We discard the traditional method (SRIE), and select SCI with the least parameters in three efficient LLIE approaches (Zero-DCE++, RUAS, and SCI) for comparison. 
Table~\ref{tab:runtime} reports the model parameters, FLOPs, and runtime averaged on 50 images of 512$\times$512. 
The running time is measured on a PC with an NVIDIA GTX 3070 GPU and an Intel i7-10700F CPU. 
Compared with the baselines, our CUE performs best with a balanced model size and runtime.

\subsection{Ablation Studies}
We conduct ablation studies on the LOL dataset to further investigate the effectiveness of our proposed CUE under different configurations.

\textbf{Effect of the number of unfolding stages:}
To explore the effect of the number of unfolding stages in the Retinex decomposition stage, we conduct experiments on proposed CUE with different numbers of stage $K$.
As illustrated in Table~\ref{tab:stageK}, the performance increases as the number of stages increases until reaching 3. 
The performance will decline when further increasing K, which may be hampered by gradient propagation. 
We set $\rm K=3$ as the default stage number to balance the performance and computation cost. 
The optimal $K$ and performance trend on the Huawei dataset is consistent with that of LOL (see supplementary material).

\begin{table}[!t]
\caption{PSNR and SSIM scores of the proposed CUE with different numbers of stages on the LOL dataset.}
\vspace{-0.1cm}
\begin{center}
\begin{center}
	\setlength{\tabcolsep}{7mm}{
		\begin{tabular}{c|cc}
			\toprule 
            Stage ($K$) & PSNR$\uparrow$ & SSIM$\uparrow$ \\ \midrule
            1  & 21.06 & 0.739 \\
            2  & 21.48 & 0.767 \\ 
            \textbf{3}  & \textbf{21.67} & \textbf{0.774} \\ 
            4  & 21.65 & 0.771\\
            5  & 21.54 & 0.766\\
			\bottomrule
	\end{tabular}}
\end{center}
\end{center}
\vspace{-0.5cm}
\label{tab:stageK}
\end{table}

\begin{figure}[t]
	\centering
	\includegraphics[width=0.48\textwidth]{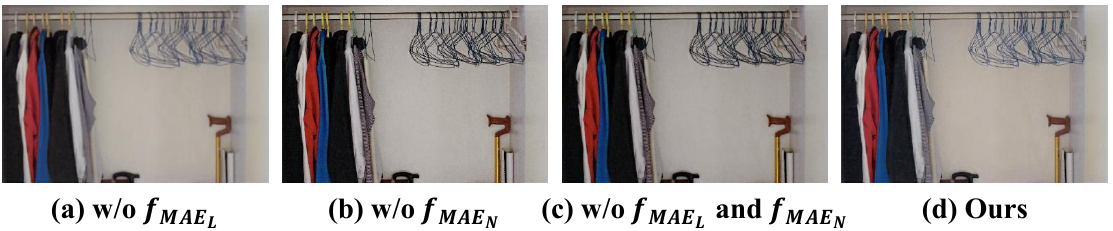}
    \vspace{-0.5cm}
	\caption{Ablation studies of $f_{MAE_{L}}$ and $f_{MAE_{N}}$.}
	\label{fig:ablation}
	\vspace{-0.6cm}
\end{figure}

\textbf{Effect of different components:}
We investigate the effectiveness of the proposed key components, i.e.,  $f_{MAE_{L}}$ and $f_{MAE_{N}}$.
The quantitative and qualitative evaluations are presented in Table~\ref{tab:component} and Fig.~\ref{fig:ablation}. 
$f_{MAE_{L}}$ denotes solving the L sub-problem with the encoder of the pre-trained illumination prior, and $f_{MAE_{N}}$ denotes employing the gradient representation prior as the regularization. 
As shown in Fig.~\ref{fig:ablation}, removing $f_{MAE_{L}}$ will cause the illumination map and the enhanced image to be over-smoothed. 
The enhanced image will suffer from severe noise pollution when discarding $f_{MAE_{N}}$.
Combining them will achieve a notable performance improvement and a satisfying visual quality. 
Supplementation presents the detailed visual results.

\subsection{Extension}
\begin{figure}[t]
	\centering
	\includegraphics[width=0.48\textwidth]{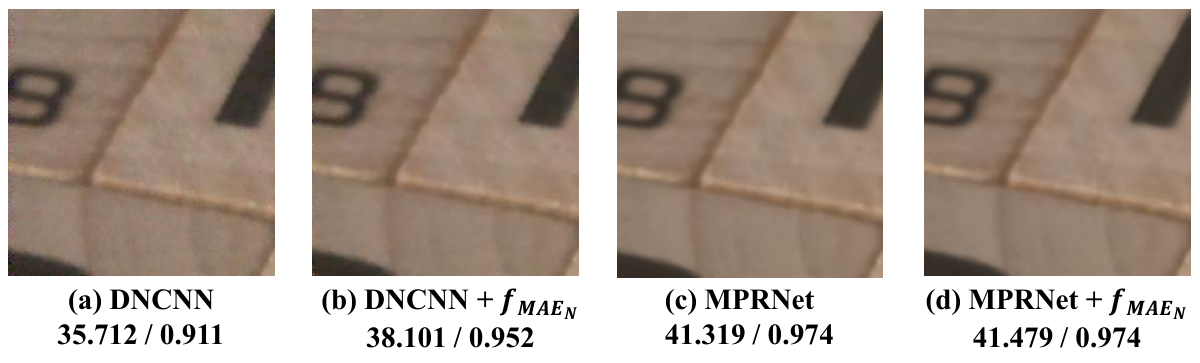}
	\caption{Extension experiments on the effectiveness of $f_{{AME}_{N}}$ for noise suppression. The numbers under each figure represent the PSNR/SSIM scores.}
	\label{fig:denoise}
\end{figure}

We further verify the effectiveness of the proposed gradient representation consistency constraint in image denoising task.
We conduct the experiments on the SIDD~\cite{SIDD} dataset with two well-known baselines, DNCNN~\cite{DNCNN} and MPRNet~\cite{MPRNet}. 
Table~\ref{tab:denoise} and Fig.~\ref{fig:denoise} demonstrate that incorporating the gradient representation consistency constraint will improve the performance of the baselines. 

\begin{table}[!t]
\caption{PSNR and SSIM scores of the ablation studies for $f_{MAE_{L}}$ and $f_{MAE_{N}}$ on the LOL dataset.}
\vspace{-0.2cm}
\begin{center}
\begin{center}
	\setlength{\tabcolsep}{4mm}{
		\begin{tabular}{cc|cc}
            \toprule
            $f_{MAE_{L}}$ & $f_{MAE_{N}}$ & PSNR$\uparrow$ & SSIM$\uparrow$ \\
             \midrule
              &   &    21.05 & 0.749 \\
            \checkmark & & 21.48 & 0.752 \\
             &  \checkmark & 21.54 & 0.758 \\
            \checkmark  & \checkmark   &\textbf{21.67} & \textbf{0.774} \\
			\bottomrule
	\end{tabular}}
\end{center}
\end{center}
\vspace{-0.5cm}
\label{tab:component}
\end{table}

\begin{table}[!t]
\caption{Extension experiments of the gradient representation consistency, $f_{MAE_{N}}$, on image denoising task. The performance is evaluated on the SIDD~\cite{SIDD} dataset.}
\vspace{-0.4cm}
\begin{center}
\begin{center}
	\setlength{\tabcolsep}{5mm}{
		\begin{tabular}{c|cc}
            \toprule
            Method & PSNR$\uparrow$ & SSIM$\uparrow$ \\
            \midrule
            DNCNN~\cite{DNCNN} &    32.43 & 0.841 \\
            DNCNN~\cite{DNCNN} + $f_{MAE_{N}}$ & \textbf{32.68} & \textbf{0.857} \\
            \midrule
            MPRNet~\cite{MPRNet} & 39.25 & 0.956 \\
            MPRNet~\cite{MPRNet} + $f_{MAE_{N}}$ & \textbf{39.32} & \textbf{0.957} \\
			\bottomrule
	\end{tabular}}
\end{center}
\end{center}
\vspace{-0.8cm}
\label{tab:denoise}
\end{table}

\section{Conclusion}
In this paper, we have demonstrated the potential of customized learnable priors for enhancing low-light images and proposed a more transparent and interpretable deep unfolding paradigm.
The customized illumination and noise priors are pre-trained using a masked image modeling strategy with specific target features. 
The learnable illumination prior is responsible for capturing the illumination properties of normal-light images, and we have embedded it into the unfolding architecture to improve its transparency in the structure flow. The noise prior aims to learn the gradient representation of normal-light images, and we have redeveloped it as a regularization term within the loss function to eliminate noise in the model output during the optimization flow.
Extensive experiments on multiple low-light image datasets have shown the superiority of our deep unfolding paradigm.

\section*{Acknowledgments}
This work was supported by the JKW Research Funds under Grant 20-163-14-LZ-001-004-01, and the Anhui Provincial Natural Science Foundation under Grant 2108085UD12. We acknowledge the support of GPU cluster built by MCC Lab of Information Science and Technology Institution, USTC.

{\small
\bibliographystyle{ieee_fullname}
\bibliography{egbib}

\begin{thebibliography}{10}\itemsep=-1pt

\bibitem{SIDD}
Abdelrahman Abdelhamed, Stephen Lin, and Michael~S Brown.
\newblock A high-quality denoising dataset for smartphone cameras.
\newblock In {\em Proceedings of the IEEE Conference on Computer Vision and
  Pattern Recognition}, pages 1692--1700, 2018.

\bibitem{histogramModification}
Tarik Arici, Salih Dikbas, and Yucel Altunbasak.
\newblock A histogram modification framework and its application for image
  contrast enhancement.
\newblock {\em IEEE Transactions on Image Processing}, 18(9):1921--1935, 2009.

\bibitem{beit}
Hangbo Bao, Li Dong, and Furu Wei.
\newblock Beit: Bert pre-training of image transformers.
\newblock {\em arXiv preprint arXiv:2106.08254}, 2021.

\bibitem{contrastEnhance}
Turgay Celik and Tardi Tjahjadi.
\newblock Contextual and variational contrast enhancement.
\newblock {\em IEEE Transactions on Image Processing}, 20(12):3431--3441, 2011.

\bibitem{bilateralGrid}
Jiawen Chen, Sylvain Paris, and Fr{\'e}do Durand.
\newblock Real-time edge-aware image processing with the bilateral grid.
\newblock {\em ACM Transactions on Graphics}, 26(3):103--es, 2007.

\bibitem{CAE}
Xiaokang Chen, Mingyu Ding, Xiaodi Wang, Ying Xin, Shentong Mo, Yunhao Wang,
  Shumin Han, Ping Luo, Gang Zeng, and Jingdong Wang.
\newblock Context autoencoder for self-supervised representation learning.
\newblock {\em arXiv preprint arXiv:2202.03026}, 2022.

\bibitem{hog}
Navneet Dalal and Bill Triggs.
\newblock Histograms of oriented gradients for human detection.
\newblock In {\em Proceedings of the IEEE Conference on Computer Vision and
  Pattern Recognition}, volume~1, pages 886--893, 2005.

\bibitem{bert}
Jacob Devlin, Ming-Wei Chang, Kenton Lee, and Kristina Toutanova.
\newblock {BERT}: Pre-training of deep bidirectional transformers for language
  understanding.
\newblock {\em arXiv preprint arXiv:1810.04805}, 2018.

\bibitem{semantic_retinex}
Minhao Fan, Wenjing Wang, Wenhan Yang, and Jiaying Liu.
\newblock Integrating semantic segmentation and {R}etinex model for low-light
  image enhancement.
\newblock In {\em Proceedings of the 28th ACM International Conference on
  Multimedia}, pages 2317--2325, 2020.

\bibitem{fu2015probabilistic}
Xueyang Fu, Yinghao Liao, Delu Zeng, Yue Huang, Xiao-Ping Zhang, and Xinghao
  Ding.
\newblock A probabilistic method for image enhancement with simultaneous
  illumination and reflectance estimation.
\newblock {\em IEEE Transactions on Image Processing}, 24(12):4965--4977, 2015.

\bibitem{SRIE}
Xueyang Fu, Delu Zeng, Yue Huang, Xiao-Ping Zhang, and Xinghao Ding.
\newblock A weighted variational model for simultaneous reflectance and
  illumination estimation.
\newblock In {\em Proceedings of the IEEE Conference on Computer Vision and
  Pattern Recognition}, pages 2782--2790, 2016.

\bibitem{ZERO_DCE}
Chunle Guo, Chongyi Li, Jichang Guo, Chen~Change Loy, Junhui Hou, Sam Kwong,
  and Runmin Cong.
\newblock Zero-reference deep curve estimation for low-light image enhancement.
\newblock In {\em Proceedings of the IEEE Conference on Computer Vision and
  Pattern Recognition}, pages 1780--1789, 2020.

\bibitem{huawei}
Jiang Hai, Zhu Xuan, Ren Yang, Yutong Hao, Fengzhu Zou, Fang Lin, and Songchen
  Han.
\newblock {R2RNet}: Low-light image enhancement via real-low to real-normal
  network.
\newblock {\em arXiv preprint arXiv:2106.14501}, 2021.

\bibitem{MAE}
Kaiming He, Xinlei Chen, Saining Xie, Yanghao Li, Piotr Doll{\'a}r, and Ross
  Girshick.
\newblock Masked autoencoders are scalable vision learners.
\newblock In {\em Proceedings of the IEEE/CVF Conference on Computer Vision and
  Pattern Recognition}, pages 16000--16009, 2022.

\bibitem{Huang2022CVPR}
Jie Huang, Yajing Liu, Xueyang Fu, Man Zhou, Yang Wang, Feng Zhao, and Zhiwei
  Xiong.
\newblock Exposure normalization and compensation for multiple-exposure
  correction.
\newblock In {\em Proceedings of the IEEE/CVF Conference on Computer Vision and
  Pattern Recognition}, pages 6043--6052, June 2022.

\bibitem{FECNet}
Jie Huang, Yajing Liu, Feng Zhao, Keyu Yan, Jinghao Zhang, Yukun Huang, Man
  Zhou, and Zhiwei Xiong.
\newblock Deep {F}ourier-based exposure correction network with
  spatial-frequency interaction.
\newblock In {\em Proceedings of the European Conference on Computer Vision},
  pages 163--180. Springer, 2022.

\bibitem{HPEU}
Jie Huang, Zhiwei Xiong, Xueyang Fu, Dong Liu, and Zheng-Jun Zha.
\newblock Hybrid image enhancement with progressive laplacian enhancing unit.
\newblock In {\em Proceedings of the 27th ACM International Conference on
  Multimedia}, page 1614–1622, 2019.

\bibitem{Huang2023CVPR}
Jie Huang, Feng Zhao, Man Zhou, Jie Xiao, Naishan Zheng, Kaiwen Zheng, and
  Zhiwei Xiong.
\newblock Learning sample relationship for exposure correction.
\newblock In {\em Proceedings of the IEEE/CVF Conference on Computer Vision and
  Pattern Recognition}, pages 9904--9913, June 2023.

\bibitem{ECLNet}
Jie Huang, Man Zhou, Yajing Liu, Mingde Yao, Feng Zhao, and Zhiwei Xiong.
\newblock Exposure-consistency representation learning for exposure correction.
\newblock In {\em Proceedings of the 30th ACM International Conference on
  Multimedia}, page 6309–6317, 2022.

\bibitem{dynamicHistogram}
Haidi Ibrahim and Nicholas Sia~Pik Kong.
\newblock Brightness preserving dynamic histogram equalization for image
  contrast enhancement.
\newblock {\em IEEE Transactions on Consumer Electronics}, 53(4):1752--1758,
  2007.

\bibitem{EnlightenGAN}
Yifan Jiang, Xinyu Gong, Ding Liu, Yu Cheng, Chen Fang, Xiaohui Shen, Jianchao
  Yang, Pan Zhou, and Zhangyang Wang.
\newblock Enlightengan: Deep light enhancement without paired supervision.
\newblock {\em IEEE Transactions on Image Processing}, 30:2340--2349, 2021.

\bibitem{jin2021bridging}
Dian Jin, Long Ma, Risheng Liu, and Xin Fan.
\newblock Bridging the gap between low-light scenes: Bilevel learning for fast
  adaptation.
\newblock In {\em Proceedings of the 29th ACM International Conference on
  Multimedia}, pages 2401--2409, 2021.

\bibitem{land1977retinex}
Edwin~H Land.
\newblock The {R}etinex theory of color vision.
\newblock {\em Scientific American}, 237(6):108--129, 1977.

\bibitem{li2021low}
Chongyi Li, Chunle Guo, Linghao Han, Jun Jiang, Ming-Ming Cheng, Jinwei Gu, and
  Chen~Change Loy.
\newblock Low-light image and video enhancement using deep learning: A survey.
\newblock {\em IEEE Transactions on Pattern Analysis and Machine Intelligence},
  44(12):9396--9416, 2021.

\bibitem{ZERO_DCE++}
Chongyi Li, Chunle Guo, and Chen~Change Loy.
\newblock Learning to enhance low-light image via zero-reference deep curve
  estimation.
\newblock {\em arXiv preprint arXiv:2103.00860}, 2021.

\bibitem{li2023embedding}
Chongyi Li, Chun-Le Guo, Man Zhou, Zhexin Liang, Shangchen Zhou, Ruicheng Feng,
  and Chen~Change Loy.
\newblock Embedding {F}ourier for ultra-high-definition low-light image
  enhancement.
\newblock In {\em International Conference on Learning Representations}, 2023.

\bibitem{LPNet}
Jiaqian Li, Juncheng Li, Faming Fang, Fang Li, and Guixu Zhang.
\newblock Luminance-aware pyramid network for low-light image enhancement.
\newblock {\em IEEE Transactions on Multimedia}, 23:3153--3165, 2020.

\bibitem{Roubust_retinex}
Mading Li, Jiaying Liu, Wenhan Yang, Xiaoyan Sun, and Zongming Guo.
\newblock Structure-revealing low-light image enhancement via robust {R}etinex
  model.
\newblock {\em IEEE Transactions on Image Processing}, 27(6):2828--2841, 2018.

\bibitem{DSLR}
Seokjae Lim and Wonjun Kim.
\newblock D{SLR}: Deep stacked laplacian restorer for low-light image
  enhancement.
\newblock {\em IEEE Transactions on Multimedia}, 23:4272--4284, 2020.

\bibitem{liu2022twin}
Risheng Liu, Zhiying Jiang, Shuzhou Yang, and Xin Fan.
\newblock Twin adversarial contrastive learning for underwater image
  enhancement and beyond.
\newblock {\em IEEE Transactions on Image Processing}, 31:4922--4936, 2022.

\bibitem{liu2022learning}
Risheng Liu, Long Ma, Tengyu Ma, Xin Fan, and Zhongxuan Luo.
\newblock Learning with nested scene modeling and cooperative architecture
  search for low-light vision.
\newblock {\em IEEE Transactions on Pattern Analysis and Machine Intelligence},
  45(5):5953--5969, 2022.

\bibitem{RUAS}
Risheng Liu, Long Ma, Jiaao Zhang, Xin Fan, and Zhongxuan Luo.
\newblock Retinex-inspired unrolling with cooperative prior architecture search
  for low-light image enhancement.
\newblock In {\em Proceedings of the IEEE Conference on Computer Vision and
  Pattern Recognition}, pages 10561--10570, 2021.

\bibitem{liu2021underexposed}
Risheng Liu, Long Ma, Yuxi Zhang, Xin Fan, and Zhongxuan Luo.
\newblock Underexposed image correction via hybrid priors navigated deep
  propagation.
\newblock {\em IEEE Transactions on Neural Networks and Learning Systems},
  33(8):3425--3436, 2021.

\bibitem{LLNet}
Kin~Gwn Lore, Adedotun Akintayo, and Soumik Sarkar.
\newblock L{LN}et: A deep autoencoder approach to natural low-light image
  enhancement.
\newblock {\em Pattern Recognition}, 61:650--662, 2017.

\bibitem{TBEFN}
Kun Lu and Lihong Zhang.
\newblock T{BEFN}: A two-branch exposure-fusion network for low-light image
  enhancement.
\newblock {\em IEEE Transactions on Multimedia}, 23:4093--4105, 2020.

\bibitem{MBLLEN}
Feifan Lv, Feng Lu, Jianhua Wu, and Chongsoon Lim.
\newblock M{BLLEN}: Low-light image/video enhancement using cnns.
\newblock In {\em Proceedings of the British Machine Vision Conference}, pages
  1--13, 2018.

\bibitem{ma2023bilevel}
Long Ma, Dian Jin, Nan An, Jinyuan Liu, Xin Fan, and Risheng Liu.
\newblock Bilevel fast scene adaptation for low-light image enhancement.
\newblock {\em arXiv preprint arXiv:2306.01343}, 2023.

\bibitem{ma2022low}
Long Ma, Risheng Liu, Yiyang Wang, Xin Fan, and Zhongxuan Luo.
\newblock Low-light image enhancement via self-reinforced retinex projection
  model.
\newblock {\em IEEE Transactions on Multimedia}, 2022.

\bibitem{ma2021learning}
Long Ma, Risheng Liu, Jiaao Zhang, Xin Fan, and Zhongxuan Luo.
\newblock Learning deep context-sensitive decomposition for low-light image
  enhancement.
\newblock {\em IEEE Transactions on Neural Networks and Learning Systems},
  33(10):5666--5680, 2021.

\bibitem{SCI}
Long Ma, Tengyu Ma, Risheng Liu, Xin Fan, and Zhongxuan Luo.
\newblock Toward fast, flexible, and robust low-light image enhancement.
\newblock In {\em Proceedings of the IEEE/CVF Conference on Computer Vision and
  Pattern Recognition (CVPR)}, pages 5637--5646, June 2022.

\bibitem{ma2022practical}
Long Ma, Tianjiao Ma, Xinwei Xue, Xin Fan, Zhongxuan Luo, and Risheng Liu.
\newblock Practical exposure correction: Great truths are always simple.
\newblock {\em arXiv preprint arXiv:2212.14245}, 2022.

\bibitem{ma2022pia}
Tengyu Ma, Long Ma, Xin Fan, Zhongxuan Luo, and Risheng Liu.
\newblock {PIA}: Parallel architecture with illumination allocator for joint
  enhancement and detection in low-light.
\newblock In {\em Proceedings of the 30th ACM International Conference on
  Multimedia}, pages 2070--2078, 2022.

\bibitem{NIQE}
Anish Mittal, Rajiv Soundararajan, and Alan~C Bovik.
\newblock Making a “completely blind” image quality analyzer.
\newblock {\em IEEE Signal Processing Letters}, 20(3):209--212, 2012.

\bibitem{unet}
Olaf Ronneberger, Philipp Fischer, and Thomas Brox.
\newblock U-net: Convolutional networks for biomedical image segmentation.
\newblock In {\em International Conference on Medical image computing and
  computer-assisted intervention}, pages 234--241. Springer, 2015.

\bibitem{gamma}
Himanshu Singh, Anil Kumar, LK Balyan, and Girish~Kumar Singh.
\newblock A novel optimally gamma corrected intensity span maximization
  approach for dark image enhancement.
\newblock In {\em 2017 22nd International Conference on Digital Signal
  Processing (DSP)}, pages 1--5. IEEE, 2017.

\bibitem{DeepUPE}
Ruixing Wang, Qing Zhang, Chi-Wing Fu, Xiaoyong Shen, Wei-Shi Zheng, and Jiaya
  Jia.
\newblock Underexposed photo enhancement using deep illumination estimation.
\newblock In {\em Proceedings of the IEEE/CVF Conference on Computer Vision and
  Pattern Recognition}, pages 6849--6857, 2019.

\bibitem{NPE}
Shuhang Wang, Jin Zheng, Hai-Miao Hu, and Bo Li.
\newblock Naturalness preserved enhancement algorithm for non-uniform
  illumination images.
\newblock {\em IEEE Transactions on Image Processing}, 22(9):3538--3548, 2013.

\bibitem{variationalGamma}
Wei Wang, Na Sun, and Michael~K Ng.
\newblock A variational gamma correction model for image contrast enhancement.
\newblock {\em Inverse Problems \& Imaging}, 13(3):461, 2019.

\bibitem{GLADNet}
Wenjing Wang, Chen Wei, Wenhan Yang, and Jiaying Liu.
\newblock {GLADN}et: Low-light enhancement network with global awareness.
\newblock In {\em 2018 13th IEEE international conference on automatic face \&
  gesture recognition (FG 2018)}, pages 751--755. IEEE, 2018.

\bibitem{progressive_retinex}
Yang Wang, Yang Cao, Zheng-Jun Zha, Jing Zhang, Zhiwei Xiong, Wei Zhang, and
  Feng Wu.
\newblock Progressive {R}etinex: Mutually reinforced illumination-noise
  perception network for low-light image enhancement.
\newblock In {\em Proceedings of the 27th ACM International Conference on
  Multimedia}, pages 2015--2023, 2019.

\bibitem{RetinexNet}
Chen Wei, Wenjing Wang, Wenhan Yang, and Jiaying Liu.
\newblock Deep {R}etinex decomposition for low-light enhancement.
\newblock {\em arXiv preprint arXiv:1808.04560}, 2018.

\bibitem{MVP}
Longhui Wei, Lingxi Xie, Wengang Zhou, Houqiang Li, and Qi Tian.
\newblock Mvp: Multimodality-guided visual pre-training.
\newblock {\em arXiv preprint arXiv:2203.05175}, 2022.

\bibitem{URetinexNet}
Wenhui Wu, Jian Weng, Pingping Zhang, Xu Wang, Wenhan Yang, and Jianmin Jiang.
\newblock {UR}etinex-net: Retinex-based deep unfolding network for low-light
  image enhancement.
\newblock In {\em Proceedings of the IEEE/CVF Conference on Computer Vision and
  Pattern Recognition (CVPR)}, pages 5901--5910, June 2022.

\bibitem{xue2022best}
Xinwei Xue, Jia He, Long Ma, Yi Wang, Xin Fan, and Risheng Liu.
\newblock Best of both worlds: See and understand clearly in the dark.
\newblock In {\em Proceedings of the 30th ACM International Conference on
  Multimedia}, pages 2154--2162, 2022.

\bibitem{DRBN}
Wenhan Yang, Shiqi Wang, Yuming Fang, Yue Wang, and Jiaying Liu.
\newblock From fidelity to perceptual quality: A semi-supervised approach for
  low-light image enhancement.
\newblock In {\em Proceedings of the IEEE/CVF conference on computer vision and
  pattern recognition}, pages 3063--3072, 2020.

\bibitem{weightedHE}
Se-Hwan Yun, Jin~Heon Kim, and Suki Kim.
\newblock Contrast enhancement using a weighted histogram equalization.
\newblock In {\em 2011 IEEE International Conference on Consumer Electronics
  (ICCE)}, pages 203--204. IEEE, 2011.

\bibitem{MPRNet}
Syed~Waqas Zamir, Aditya Arora, Salman Khan, Munawar Hayat, Fahad~Shahbaz Khan,
  Ming-Hsuan Yang, and Ling Shao.
\newblock Multi-stage progressive image restoration.
\newblock In {\em Proceedings of the IEEE/CVF conference on computer vision and
  pattern recognition}, pages 14821--14831, 2021.

\bibitem{LowLightZhang}
Jinghao Zhang, Jie Huang, Mingde Yao, Man Zhou, and Feng Zhao.
\newblock Structure- and texture-aware learning for low-light image
  enhancement.
\newblock In {\em Proceedings of the 30th ACM International Conference on
  Multimedia}, page 6483–6492, 2022.

\bibitem{DNCNN}
Kai Zhang, Wangmeng Zuo, Yunjin Chen, Deyu Meng, and Lei Zhang.
\newblock Beyond a gaussian denoiser: Residual learning of deep cnn for image
  denoising.
\newblock {\em IEEE transactions on Image Processing}, 26(7):3142--3155, 2017.

\bibitem{KinD++}
Yonghua Zhang, Xiaojie Guo, Jiayi Ma, Wei Liu, and Jiawan Zhang.
\newblock Beyond brightening low-light images.
\newblock {\em International Journal of Computer Vision}, 129(4):1013--1037,
  2021.

\bibitem{KinD}
Yonghua Zhang, Jiawan Zhang, and Xiaojie Guo.
\newblock Kindling the darkness: A practical low-light image enhancer.
\newblock In {\em Proceedings of the ACM International Conference on
  Multimedia}, pages 1632--1640, 2019.

\bibitem{RetinexDIP}
Zunjin Zhao, Bangshu Xiong, Lei Wang, Qiaofeng Ou, Lei Yu, and Fa Kuang.
\newblock Retinex{DIP}: A unified deep framework for low-light image
  enhancement.
\newblock {\em IEEE Transactions on Circuits and Systems for Video Technology},
  pages 1--14, 2021.

\bibitem{TVUnfolding}
Chuanjun Zheng, Daming Shi, and Wentian Shi.
\newblock Adaptive unfolding total variation network for low-light image
  enhancement.
\newblock In {\em Proceedings of the IEEE/CVF International Conference on
  Computer Vision}, pages 4439--4448, 2021.

\bibitem{9847530}
Naishan Zheng, Jie Huang, Feng Zhao, Xueyang Fu, and Feng Wu.
\newblock Unsupervised underexposed image enhancement via self-illuminated and
  perceptual guidance.
\newblock {\em IEEE Transactions on Multimedia}, pages 1--16, 2022.

\bibitem{zheng2022enhancement}
Naishan Zheng, Jie Huang, Qi Zhu, Man Zhou, Feng Zhao, and Zheng-Jun Zha.
\newblock Enhancement by your aesthetic: An intelligible unsupervised
  personalized enhancer for low-light images.
\newblock In {\em Proceedings of the 30th ACM International Conference on
  Multimedia}, pages 6521--6529, 2022.

\bibitem{zhou2023fourmer}
Man Zhou, Jie Huang, Chun-Le Guo, and Chongyi Li.
\newblock Fourmer: An efficient global modeling paradigm for image restoration.
\newblock In {\em International Conference on Machine Learning (ICML)-Oral},
  pages 42589--42601. PMLR, 2023.

\bibitem{zhou2022lednet}
Shangchen Zhou, Chongyi Li, and Chen~Change Loy.
\newblock Lednet: Joint low-light enhancement and deblurring in the dark.
\newblock In {\em European Conference on Computer Vision (ECCV)}, 2022.

\bibitem{zhuang2021bayesian}
Peixian Zhuang, Chongyi Li, and Jiamin Wu.
\newblock Bayesian retinex underwater image enhancement.
\newblock {\em Engineering Applications of Artificial Intelligence [Paper Prize
  Award 2023-Application]}, 101:104171, 2021.

\end{thebibliography}
}

\end{document}